\documentclass[conference]{IEEEtran}
\IEEEoverridecommandlockouts
\usepackage{cite}
\usepackage{amsmath,amssymb,amsfonts}
\usepackage{algorithmic}
\usepackage{graphicx}
\usepackage{textcomp}
\usepackage{xcolor}
\usepackage{booktabs}
\usepackage{multirow}

\def\BibTeX{{\rm B\kern-.05em{\sc i\kern-.025em b}\kern-.08em
    T\kern-.1667em\lower.7ex\hbox{E}\kern-.125emX}}
\usepackage{bibunits}

\defaultbibliographystyle{IEEEbib}
\defaultbibliography{icme2026references}
\begin{document}

\begin{bibunit}

\title{Modality-Specific Hierarchical Enhancement for RGB-D Camouflaged Object Detection

\thanks{* is the corresponding author. }
}

\author{
Yuzhen Niu$^{1}$, 
Yangqing Wang$^{1}$, 
Ri Cheng$^{1,*}$, 
Fusheng Li$^{1}$, 
Rongshen Wang$^{1}$, 
Zhichen Yang$^{1}$\\
$^{1}$College of Computer and Data Science, Fuzhou University, Fuzhou, China\\
yuzhenniu@gmail.com, yangqwang@163.com, rcheng22@m.fudan.edu.cn, lifusheng.chn@gmail.com,\\
w1911930384@gmail.com, zhichenyang47@gmail.com
}

\maketitle

\begin{abstract}

Camouflaged object detection (COD) is challenging due to high target-background similarity, and recent methods address this by complementarily using RGB-D texture and geometry cues. However, RGB-D COD methods still underutilize modality-specific cues, which limits fusion quality. We believe this is because RGB and depth features are fused directly after backbone extraction without modality-specific enhancement. To address this limitation, we propose MHENet, an RGB-D COD framework that performs modality-specific hierarchical enhancement and adaptive fusion of RGB and depth features. 
Specifically, we introduce a Texture Hierarchical Enhancement Module (THEM) to amplify subtle texture variations by extracting high-frequency information and a Geometry Hierarchical Enhancement Module (GHEM) to enhance geometric structures via learnable gradient extraction, while preserving cross-scale semantic consistency. Finally, an Adaptive Dynamic Fusion Module (ADFM) adaptively fuses the enhanced texture and geometry features with spatially varying weights. Experiments on four benchmarks demonstrate that MHENet surpasses 16 state-of-the-art methods qualitatively and quantitatively. Code is available at https://github.com/afdsgh/MHENet.
\end{abstract}

\begin{IEEEkeywords}
Camouflaged object detection, Hierarchical enhancement, Cross-Modal fusion
\end{IEEEkeywords}

\section{Introduction}
\label{sec:intro}

Camouflage, a survival strategy in nature, allows organisms to visually blend with their surroundings by adjusting appearance attributes such as color, texture, and shape~\cite{price2019background}. 
Similar camouflage effects also arise beyond biology, influencing human practices in art, culture, and modern visual technologies~\cite{stevens2009animal}. 
Inspired by this phenomenon, camouflaged object detection (COD)\cite{fan2020camouflaged} focuses on discovering concealed targets in complex scenes, and has found practical use in industrial defect inspection\cite{cui2021sddnet}, agricultural pest monitoring~\cite{rustia2020application}, medical lesion segmentation~\cite{fan2020pranet}, and wildlife conservation~\cite{perez2012early}. 
Nevertheless, COD remains difficult because camouflaged objects often exhibit extremely low contrast and high visual similarity to their backgrounds, making the target boundaries and regions hard to separate for both humans and learning-based models. Therefore, developing robust and accurate COD models is crucial for reliable deployment in real-world scenarios.
\begin{figure}[h]
\centering
\includegraphics[width=\linewidth]{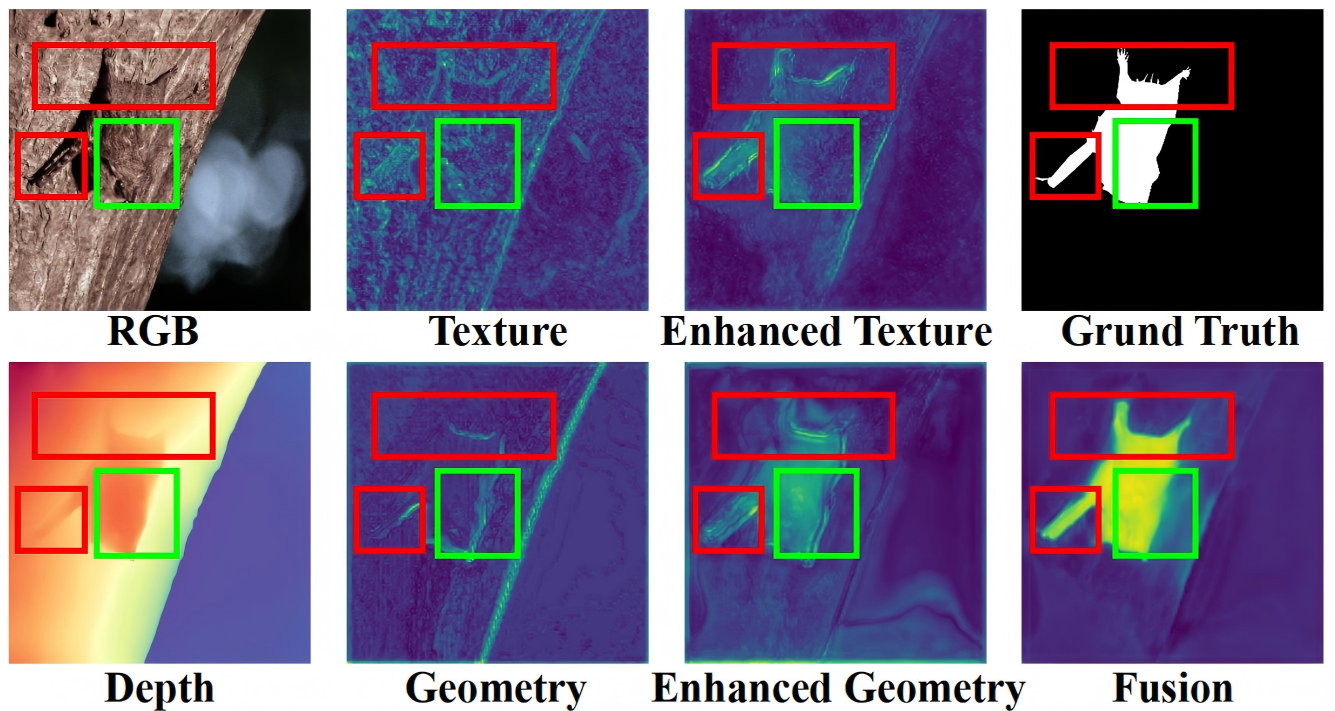}
\caption{Texture enhancement enriches the texture details of the limbs (red boxes) and geometry enhancement strengthens the bat and its boundary activations (green boxes), enabling fusion to better combine complementary cues. RGB activates more on texture-rich limbs but less on the camouflaged bat, while depth complements the bat for complementary fusion.}
\label{fig:motivation}
\vspace{-2mm}
\end{figure}

Many COD approaches operate on RGB inputs and enhance appearance cues from different perspectives, such as boundary or region modeling \cite{sun2022boundary,yue2024progressive}, multi-scale aggregation \cite{pang2022zoom}, and distraction suppression \cite{lyu2025distraction}. 
Despite these advances, RGB signals are inherently ambiguous under heavy camouflage, where texture and color cues are weak or misleading. 
Therefore, RGB-D COD introduces depth maps to complement RGB appearance, since depth offers geometric layout and spatial continuity cues that are less affected by texture camouflage.
Along this line, DaCOD~\cite{wang2023depth} introduces multi-modal collaborative learning with asymmetric fusion, RISNet~\cite{wang2024depth} integrates RGB-D features with multi-scale receptive fields and iterative refinement for challenging agricultural scenes, and CPNet~\cite{hu2024cross} adopts a dual-stream Swin Transformer with cross-modal attention and progressive decoding to refine representations.

\begin{figure*}[!t]
\centering
\includegraphics[width=\linewidth]{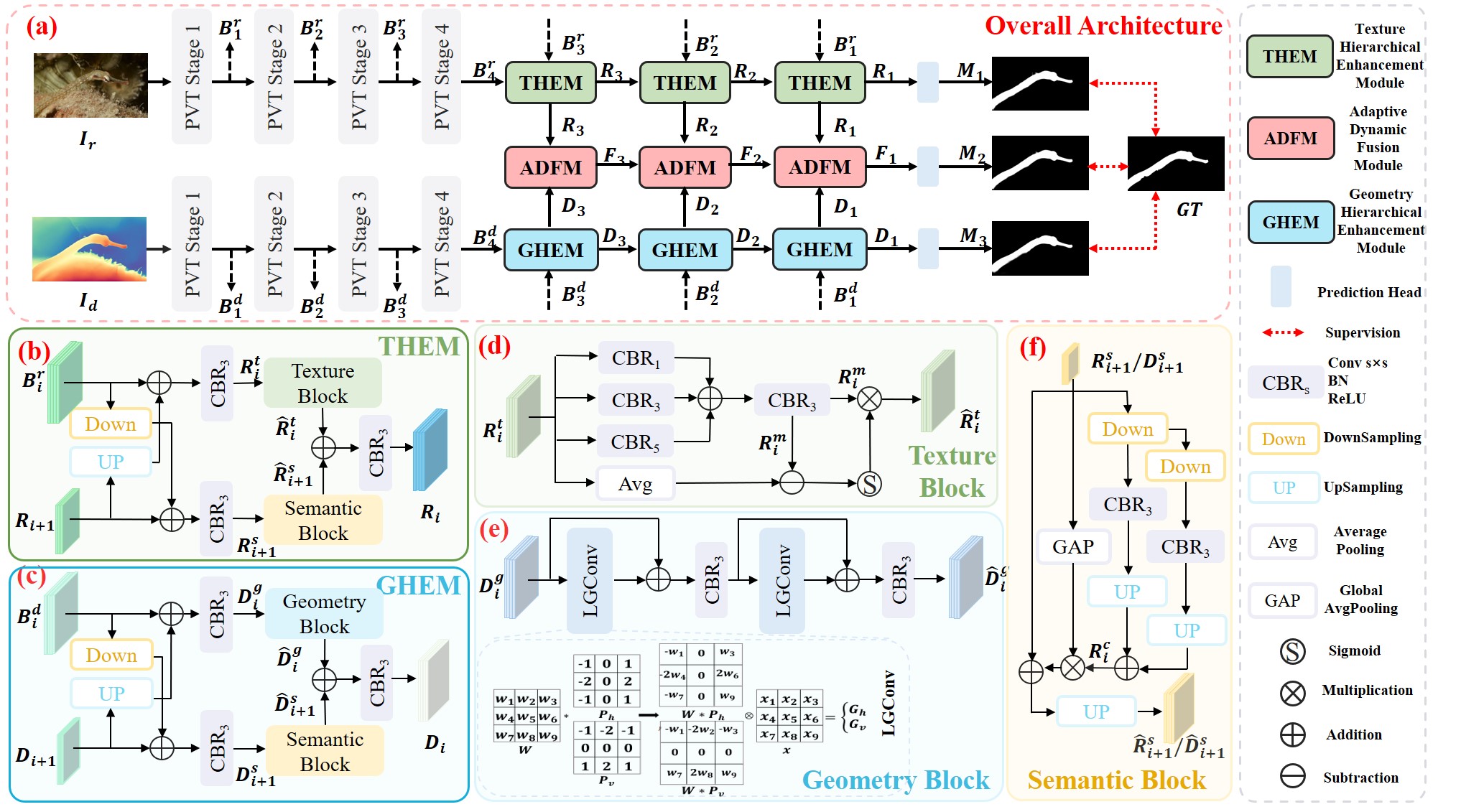}
\vspace{-2mm}
\caption{The overall architecture of the proposed MHENet, which consists of three key components, Texture Hierarchical Enhancement Module (THEM), Geometry Hierarchical Enhancement Module (GHEM), and Adaptive Dynamic Fusion Module (ADFM).}
\label{fig:architecture}
\vspace{-2mm}
\end{figure*}

Even with recent advances in RGB and RGB-D COD, camouflaged objects are still challenging to separate from intricate backgrounds.
Existing RGB-D COD methods directly fuse RGB and depth features after backbone extraction, without modality-specific enhancement, which tends to weaken texture or geometry cues and leads to suboptimal fusion.
RGB and depth features exhibit different characteristics. Specifically, RGB emphasizes fine-grained texture variations for subtle appearance discrimination \cite{galun2003texture}, whereas depth encodes geometry and spatial structure for texture-independent foreground-background separation and reliable localization \cite{wu2023source}. 
Therefore, applying the same modeling strategy to RGB and depth features, or relying only on late-stage fusion, limits the network’s ability to utilize complementary texture and geometric cues.
This observation underscores the need for modality-specific enhancement of RGB and depth features, maximizing the exploitation of their respective texture and geometric cues.

To address these limitations, we propose MHENet, a modality-aware RGB-D COD framework that explicitly enhances texture and geometric cues through hierarchical enhancement.
Specifically, we propose a Texture Hierarchical Enhancement Module (THEM) and a Geometry Hierarchical Enhancement Module (GHEM) to perform modality-specific hierarchical enhancement for RGB and depth features, respectively. 
THEM and GHEM adopt an identical hierarchical interaction framework and differ in the modality-specific enhancement block.
A shared semantic block enforces cross-scale semantic consistency by aligning high-resolution features with semantically richer low-resolution representations, facilitating enhanced information propagation across scales.
Meanwhile, THEM strengthens texture cues via high-frequency modulation in texture block, while GHEM enhances geometric structures through learnable gradient extraction in geometry block, thereby explicitly preserving modality-specific characteristics prior to fusion.
Then, an Adaptive Dynamic Fusion Module (ADFM) is further employed to selectively integrate RGB and depth representations via spatially adaptive weighting.

As shown in Fig.~\ref{fig:motivation}, texture enhancement highlights discriminative details such as the limbs (red boxes), but struggles in heavily camouflaged regions like the bat (green boxes). 
The enhanced geometry feature provides more reliable geometric cues in these areas, and spatially adaptive fusion combines both to produce more accurate camouflaged object masks.
Compared with existing methods that rely on uniform fusion strategies, MHENet explicitly accounts for both modality-specific characteristics, leading to more discriminative and robust representations for camouflaged object detection.
The main contributions of this work can be summarized as follows:



\begin{itemize}
\item We propose MHENet, a modality-aware RGB-D COD framework that explicitly enhances texture and geometric cues in a hierarchical manner, enabling more reliable detection under severe target-background similarity.

\item We propose two modality-specific hierarchical enhancement modules, THEM for RGB and GHEM for depth. Both adopt the same cross-scale alignment with a unified semantic block to maintain semantic consistency, while using a texture or a geometry block to enhance RGB texture details and depth geometric structures before the adaptive RGB-D fusion via the proposed ADFM.

\item Extensive experiments on four public benchmarks validate the effectiveness of the proposed modality-specific enhancement, and show that MHENet outperforms 4 RGB-D and 12 RGB state-of-the-art COD methods.

\end{itemize}


\section{THE PROPOSED METHOD}

\subsection{Overall Architecture}
We display the proposed MHENet in  Fig.~\ref{fig:architecture}.
Given an RGB image $I_{r}$ and its corresponding depth map $I_d$ obtained by a monocular depth estimation model \cite{he2025distill}, we adopt a dual-stream PVT~\cite{wang2021pvtv2} backbone to extract four-level multi-scale feature maps $\{B_i^r\}_{i=1}^4$ and $\{B_i^d\}_{i=1}^4$ for the RGB and depth branches.
For the RGB branch, features are progressively fed into THEM for cross-scale interaction and texture enhancement, generating three enhanced RGB features $\{R_i\}_{i=1}^3$. 
In parallel, the depth branch uses Geometry Hierarchical Enhancement Module (GHEM) to enhance geometric cues with hierarchical semantic consistency, generating $\{D_i\}_{i=1}^3$.
Then, Adaptive Dynamic Fusion Module (ADFM) is applied at each level to fuse the enhanced RGB and depth features, producing $\{F_i\}_{i=1}^3$.
$R_1$, $F_1$, and $D_1$ are fed into three prediction heads to obtain $\{M_i\}_{i=1}^3$, where $M_2$ is used as the final output. For each prediction $M_i$ $(i=1,2,3)$, we adopt the BCE loss and IoU loss as the loss functions, and more details about loss functions are provided in the supplementary material.

\subsection{Texture Hierarchical Enhancement Module (THEM)} 
In camouflaged object detection, low-level RGB features preserve fine-grained texture details for distinguishing targets from similar backgrounds, while high-level features provide semantic context for robust localization and background suppression.
To exploit these complementary cues, we propose THEM (Fig.~\ref{fig:architecture}(b)) to perform hierarchical cross-scale interaction, progressively injecting high-level semantic guidance into low-level features while enhancing discriminative texture details for accurate camouflage perception.


\noindent
\textbf{Cross-scale Alignment.}
For each pair of RGB features $\left(R_{i+1}, B_i^r\right)$ with $i\in\{1,2,3\}$, where $R_4$ is $B_4^r$, THEM aligns their spatial resolutions to enable interaction.
Specifically, the higher-level feature $R_{i+1}$ is upsampled to the resolution of $B_i^r$, while $B_i^r$ is downsampled by a factor of $2$ to match $R_{i+1}$.
Accordingly, we construct two intermediate representations:
\begin{equation}
\begin{cases}
R_i^t = {CBR}_{3\times3}(B_i^r + \mathcal{U}(R_{i+1},2)), \\
R_{i+1}^{\,s}  = {CBR}_{3\times3}(R_{i+1} + \mathcal{D}(B_i^r,2)),
\end{cases}
\label{equ.top_down}
\end{equation}
where $\mathcal{D}(\cdot,s)$ and $\mathcal{U}(\cdot,s)$ denote $s$-fold downsampling and upsampling for cross-scale resolution alignment, respectively. $\mathrm{CBR}_{3\times3}$ indicates a $3\times3$ convolution followed by batch normalization and ReLU.
The resulting $R_i^t$ and $R_{i+1}^{\,s}$ are then fed into the Texture Block and Semantic Block, respectively.

\begin{table*}[t]
\centering
\caption{Comparison of our proposed method with 16 state-of-the-art methods on four benchmark datasets across four metrics: $S_\alpha \uparrow $, $E_\varphi \uparrow$, $F_\beta^\omega \uparrow$, $M\downarrow$. The best two results are highlighted in \textcolor{red}{red} and \textcolor{blue}{blue}.}
\vspace{-2mm}
\resizebox{\linewidth}{!}{ 
\begin{tabular}{l l c c c c c c c c c c c c c c c c}
\toprule
\multirow{2}{*}{Methods} & \multirow{2}{*}{Pub.Year} & \multicolumn{4}{c}{CHAMELEON} & \multicolumn{4}{c}{CAMO-Test} & \multicolumn{4}{c}{COD10K-Test} & \multicolumn{4}{c}{NC4K} \\
\cmidrule(lr){3-6} \cmidrule(lr){7-10} \cmidrule(lr){11-14} \cmidrule(lr){15-18}
& & $S_\alpha \uparrow$ & $E_\varphi \uparrow$ & $F_\beta^\omega \uparrow$ & $M \downarrow$ & $S_\alpha \uparrow$ & $E_\varphi \uparrow$ & $F_\beta^\omega \uparrow$ & $M \downarrow$ & $S_\alpha \uparrow$ & $E_\varphi \uparrow$ & $F_\beta^\omega \uparrow$ & $M \downarrow$ & $S_\alpha \uparrow$ & $E_\varphi \uparrow$ & $F_\beta^\omega \uparrow$ & $M \downarrow$ \\
\midrule
\multicolumn{18}{c}{RGB-based COD Methods} \\
\midrule
SINet\cite{fan2020camouflaged} & CVPR.20 & 0.872 & 0.936 & 0.806 & 0.034 & 0.745 & 0.804 & 0.644 & 0.092 & 0.776 & 0.864 & 0.631 & 0.043 & 0.808 & 0.871 & 0.723 & 0.058 \\
BGNet\cite{sun2022boundary} & IJCAI.22 & 0.901 & 0.943 & 0.850 & 0.027 & 0.812 & 0.857 & 0.749 & 0.073 & 0.831 & 0.901 & 0.722 & 0.033 & 0.851 & 0.907 & 0.788 & 0.044 \\
ZoomNet\cite{pang2022zoom} & CVPR.22 & 0.902 & 0.943 & 0.845 & 0.023 & 0.820 & 0.877 & 0.752 & 0.066 & 0.838 & 0.888 & 0.729 & 0.029 & 0.853 & 0.896 & 0.784 & 0.043 \\
FSPNet\cite{huang2023feature} & CVPR.23 & 0.908 & 0.943 & 0.851 & 0.023 & 0.856 & 0.899 & 0.799 & 0.050 & 0.851 & 0.895 & 0.735 & 0.026 & 0.879 & 0.915 & 0.816 & 0.035 \\
UEDG\cite{lyu2023uedg} & TMM.23 & 0.911 & 0.958 & 0.866 & 0.023 & 0.863 & 0.922 & 0.817 & 0.048 & 0.858 & 0.924 & 0.766 & 0.025 & 0.879 & 0.929 & 0.830 & 0.035 \\
MSCAF\cite{liu2023mscaf} & TCSVT.23 & 0.912 & 0.958 & 0.865 & 0.022 & 0.873 & \textcolor{blue}{0.929} & 0.828 & 0.046 & 0.865 & 0.927 & 0.775 & 0.024 & 0.887 & \textcolor{blue}{0.934} & 0.838 & \textcolor{blue}{0.032} \\
DINet\cite{zhou2024decoupling} & TMM.24 & - & - & - & - & 0.821 & 0.874 & 0.790 & 0.068 & 0.832 & 0.903 & 0.761 & 0.031 & 0.856 & 0.909 & 0.824 & 0.043 \\
RISNet\cite{wang2024depth} & CVPR.24 & - & - & - & - & 0.870 & 0.922 & 0.827 & 0.050 & \textcolor{blue}{0.873} & 0.931 & \textcolor{blue}{0.799} & 0.025 & 0.882 & 0.925 & 0.834 & 0.037 \\
ICEG\cite{he2023strategic} & ICLR.24 & 0.905 & \textcolor{blue}{0.959} & 0.860 & 0.023 & 0.867 & 0.926 & 0.855 & \textcolor{blue}{0.044} & 0.857 & 0.930 & 0.782 & 0.024 & 0.879 & 0.932 & \textcolor{red}{0.855} & 0.034 \\
DSNet\cite{lyu2025distraction} & ICME.25 & \textcolor{blue}{0.914} & - & 0.867 & 0.022 & 0.868 & - & 0.826 & 0.048 & 0.867 & - & 0.783 & \textcolor{blue}{0.023} & 0.884 & - & 0.839 & \textcolor{red}{0.031} \\
PRBENet\cite{yue2024progressive} & TMM.25 & \textcolor{red}{0.918} & 0.951 & \textcolor{red}{0.878} & \textcolor{blue}{0.020} & 0.876 & 0.928 & 0.837 & 0.045 & 0.867 & \textcolor{blue}{0.932} & 0.793 & \textcolor{blue}{0.023} & 0.887 & 0.931 & 0.845 & \textcolor{red}{0.031} \\
SENet\cite{hao2025simple} & TIP.25 & \textcolor{red}{0.918} & 0.957 & \textcolor{red}{0.878} & \textcolor{red}{0.019} & \textcolor{red}{0.888} & \textcolor{red}{0.932} & \textcolor{red}{0.847} & \textcolor{red}{0.039} & 0.865 & 0.925 & 0.780 & 0.024 & \textcolor{blue}{0.889} & 0.933 & 0.843 & \textcolor{blue}{0.032} \\
MHENet (Ours) & - & \textcolor{blue}{0.914} & \textcolor{red}{0.966} & \textcolor{blue}{0.872} & 0.022 & \textcolor{blue}{0.883} & \textcolor{red}{0.932} & \textcolor{blue}{0.840} & 0.045 & \textcolor{red}{0.880} & \textcolor{red}{0.937} & \textcolor{red}{0.803} & \textcolor{red}{0.021} & \textcolor{red}{0.895} & \textcolor{red}{0.936} & \textcolor{blue}{0.851} & \textcolor{red}{0.031} \\
\midrule
\multicolumn{18}{c}{RGB-D-based COD Methods} \\
\midrule

DaCOD\cite{wang2023depth} & MM.23 & - & - & - & - & 0.855 & 0.911 & 0.796 & 0.051 & 0.840 & 0.908 & 0.729 & 0.028 & 0.874 & 0.923 & 0.814 & 0.035 \\
DSAM\cite{yu2024exploring} & MM.24 & - & - & - & - & 0.832 & - & 0.794 & 0.061 & 0.846 & - & 0.760 & 0.033 & 0.871 & - & 0.826 & 0.040 \\
MAGNet\cite{zhong2024magnet} & KBS.24 & 0.917 & \textcolor{blue}{0.963} & 0.876 & 0.019 & \textcolor{blue}{0.888} & \textcolor{red}{0.933} & \textcolor{blue}{0.848} & \textcolor{red}{0.037} & 0.868 & \textcolor{blue}{0.929} & \textcolor{blue}{0.792} & 0.025 & 0.886 & \textcolor{blue}{0.931} & \textcolor{blue}{0.841} & 0.033 \\
MultiCOS\cite{fang2025integrating} & Arxiv.25 & \textcolor{blue}{0.923} & - & - & 0.018 & 0.867 & - & - & 0.048 & \textcolor{blue}{0.880} & - & - & \textcolor{blue}{0.020} & \textcolor{blue}{0.890} & - & - & \textcolor{blue}{0.031} \\
MHENet (Ours) & - & \textcolor{red}{0.926} & \textcolor{red}{0.966} & \textcolor{red}{0.891} & \textcolor{red}{0.018} & \textcolor{red}{0.893} & \textcolor{blue}{0.932} & \textcolor{red}{0.852} & \textcolor{blue}{0.038} & \textcolor{red}{0.889} & \textcolor{red}{0.942} & \textcolor{red}{0.817} & \textcolor{red}{0.019} & \textcolor{red}{0.902} & \textcolor{red}{0.939} & \textcolor{red}{0.859} & \textcolor{red}{0.029} \\
\bottomrule
\end{tabular}
}

\label{tab:comparison}
\end{table*}

\noindent
\textbf{Texture Block.}
The Texture Block is designed to highlight subtle texture discrepancies between camouflaged objects and their surrounding backgrounds, which are often indistinguishable in terms of overall color and appearance.
As shown in Fig.~\ref{fig:architecture}(d), multi-scale texture cues are extracted from $R_i^t$ using parallel convolutions with different kernel sizes, producing $R_i^m$ for camouflage separation.
To suppress low-frequency components and emphasize local texture contrast, we subtract the average-pooled aligned feature from the aggregated texture feature.
Finally, a sigmoid gating is applied to enhance texture-salient regions.
The texture-enhanced feature is computed as:
\begin{equation}
\begin{cases}
\begin{aligned}
R^m_i &= CBR_{3 \times 3}(\sum_{k \in \{1,3,5\}} CBR_{k \times k}(R_i^t)), \\
\hat{R}_i^{t} &= \sigma\left( R^m_i - AVG(R_i^t) \right) \otimes R^m_i,
\end{aligned}
\end{cases}
\end{equation}
where $\sigma(\cdot)$ denotes the sigmoid function and $\otimes$ represents element-wise multiplication, $AVG(\cdot)$ is average pooling. 
This operation adaptively highlights informative texture regions.

\noindent
\textbf{Semantic Block.}
As shown in Fig.~\ref{fig:architecture}(f), the Semantic Block is designed to strengthen and propagate high-level semantic guidance to lower-level features after cross-scale alignment (Eq.~\ref{equ.top_down}), ensuring semantic consistency across scales for more reliable camouflage separation.
Given the semantic feature $R^s_{i+1}$, we first build multi-scale contextual features by cascading downsampling and $3{\times}3$ convolutions, and then upsample and merge them to form a context-enhanced semantic feature $R^c_i$.
We further compute a global channel descriptor by $GAP(R^s_{i+1})$ to gate $R^c_i$, adaptively reweighting its channels with global semantic guidance.
Finally, we add the gated context back to $R^s_{i+1}$ via a residual connection, and upsample the result to obtain $\hat{R}^s_{i+1}$:
\begin{equation}
\begin{cases}
R^s_{i+1,1}=CBR_{3\times3}(\mathcal{D}(R^s_{i+1},2)),\\
R^s_{i+1,2}=CBR_{3\times3}(\mathcal{D}(R^s_{i+1,1},2)),\\
R^c_i = \mathcal{U}(R^s_{i+1,1},2)+\mathcal{U}(R^s_{i+1,2},4),\\
\hat{R}^s_{i+1} = \mathcal{U}((R^c_i\otimes GAP(R^s_{i+1}) + R^s_{i+1}),2),
\end{cases}
\label{equ:semantic_block}
\end{equation}
where $GAP(\cdot)$ denotes global average pooling.


\noindent
\textbf{Texture-Semantic Fusion.}
The enhanced texture feature $\hat{R}^t_{i}$ is fused with the semantic feature $\hat{R}^s_{i+1}$ as follows:
\begin{equation}
R_i = CBR_{3\times 3}(\hat{R}^t_{i} + \hat{R}^s_{i+1}).
\end{equation}
The resulting feature map $R_i$ encodes enhanced texture details while preserving cross-scale semantic consistency and enriched texture cues, producing an enhanced RGB representation for subsequent cross-modal fusion.

\subsection{Geometry Hierarchical Enhancement Module (GHEM)} 
Compared with RGB features that mainly depend on subtle texture differences and semantic understanding, depth features offer geometry and structure cues that are less affected by appearance camouflage, thereby supporting texture-independent separation and reliable localization.
To fully exploit geometric cues, we propose a Geometry Hierarchical Enhancement Module (GHEM) to refine depth features, as shown in Fig.~\ref{fig:architecture}(c).
Given two depth features $D_{i+1}$ and $B_i^d$, GHEM follows the same cross-scale alignment strategy as THEM (Eq.~\ref{equ.top_down}) to obtain aligned depth features $D_i^g$ and $D_{i+1}^s$, which are fed into the Geometry Block and the Semantic Block, respectively.

\noindent
\textbf{Geometry Block.}
The Geometry Block is designed to explicitly enhance structural variations in depth features by modeling local geometric gradients, which are crucial for distinguishing camouflaged objects from cluttered backgrounds.
To this end, as shown in Fig.~\ref{fig:architecture}(e), we introduce a learnable gradient convolution $LGConv(\cdot)$, which computes horizontal and vertical gradients using learnable convolution kernels initialized with Sobel operators.
Given an input feature map $x \in \mathbb{R}^{C \times H \times W}$, we define two basic matrices $P_h$ and $P_v$ corresponding to the horizontal and vertical Sobel operators, respectively. We then modulate these basic matrices with the learnable convolution kernel $W$ via element-wise multiplication to obtain adaptive gradient kernels, which are convolved with $x$. The process of $LGConv(\cdot)$ is summarized below:
\begin{equation}
LGConv(x)=
\begin{cases}
G_h=x\otimes(W* P_h),\\ 
G_v=x\otimes(W* P_v), \\
G = \sqrt{G_h^2 + G_v^2 + \epsilon},
\end{cases}
\label{equ:G}
\end{equation}
where $G$ is the combined magnitude and $\epsilon$ is a small constant for numerical stability.


Built upon $LGConv(\cdot)$, the Geometry Block enhances geometric cues in depth features via two successive gradient refinements:
\begin{equation}
\begin{cases}
D^g = CBR_{3\times 3}(LGConv(D_i^g) + D_i^g),\\
\hat{D}_i^g = CBR_{3\times 3}(LGConv(D^g) + D^g),
\end{cases}
\end{equation}
This process strengthens reliable structures and suppresses depth noise.

\begin{figure}[!t]
\centering
\includegraphics[width=0.95\linewidth]{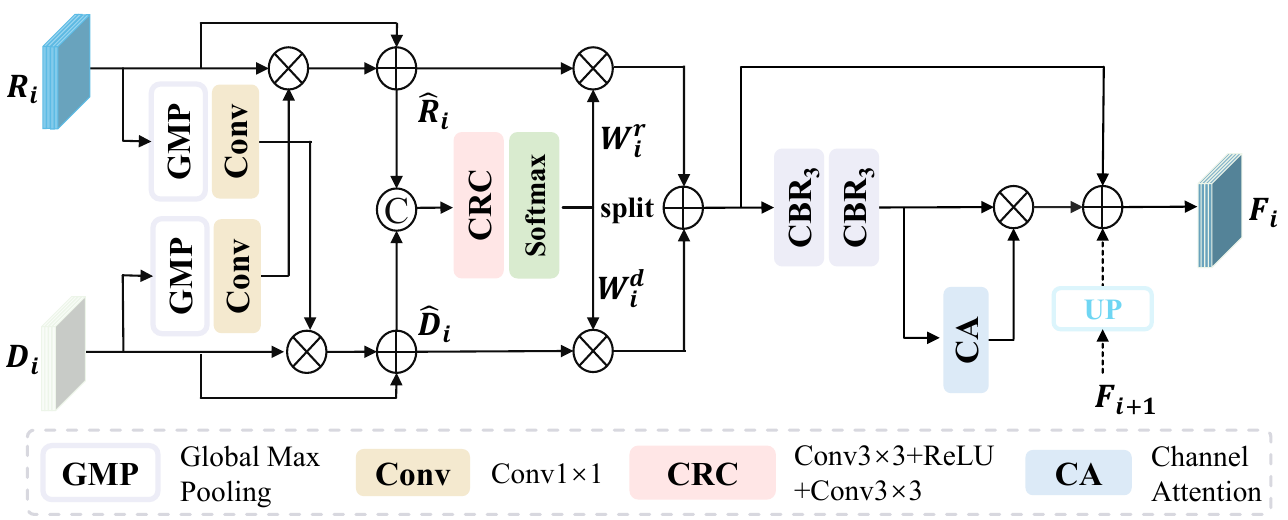}
\caption{Overview of the Adaptive Dynamic Fusion Module.}
\label{fig:adfm}
\end{figure}

\noindent
\textbf{Geometry-Semantic Fusion.}
The semantic base feature $D_{i+1}^s$ is fed into the Semantic Block (Eq.~\ref{equ:semantic_block}) to facilitate cross-scale geometric semantic consistency, producing $\hat{D}_{i+1}^s$.
Finally, we fuse the geometry-enhanced feature $\hat{D}_i^g$ with the semantic feature $\hat{D}_{i+1}^s$ to obtain the output of GHEM:
\begin{equation}
D_i = CBR_{3\times 3}\big(\hat{D}_i^g + \hat{D}_{i+1}^s\big).
\end{equation}
The resulting feature $D_i$ integrates geometry-aware structure enhancement with semantically consistent context, providing a robust depth representation for subsequent cross-modal fusion.

\subsection{Adaptive Dynamic Fusion Module (ADFM)} 
The Adaptive Dynamic Fusion Module (ADFM) aims to adaptively fuse enhanced RGB and depth features by exploiting cross-modal guidance and spatially varying modality selection, so that reliable cues are emphasized while conflicting cues are suppressed. Given $R_i$ and $D_i$ at scale $i$, ADFM first performs cross-modal global guidance and then conducts gated fusion:
\begin{align}
\left\{
\begin{aligned}
&\hat{R}_i = R_i \otimes {Conv}_{1\times1}\!\big({GMP}(D_i)\big) + R_i,\\
&\hat{D}_i = D_i \otimes {Conv}_{1\times1}\!\big({GMP}(R_i)\big) + D_i,\\
&W_i^r,\,W_i^d = {Split}\!\Big( \mathrm{Softmax}\big(CRC([\hat{R}_i,\hat{D}_i])\big) \Big),\\
&F_i^{m} = W_i^r \otimes \hat{R}_i + W_i^d \otimes \hat{D}_i,
\end{aligned}
\right.
\tag{8}
\end{align}
where ${GMP}(\cdot)$ denotes global max pooling, ${Conv}_{1\times1}(\cdot)$ is a $1\times1$ convolution, $[\cdot,\cdot]$ represents channel-wise concatenation, and ${Split}(\cdot)$ splits the predicted weights along the channel direction.
The spatial weights $W_i^r$ and $W_i^d$ are predicted from $[\hat{R}_i,\hat{D}_i]$ and normalized by Softmax along the modality dimension, leading to $W_i^r+W_i^d=1$ at each spatial location.

Subsequently, the fused feature $F_i^{m}$ is refined and integrated with cross-scale context. Specifically, we apply two consecutive $CBR$ blocks to obtain a transformed feature $F_{\text{ref}}$, and then use the channel attention module to generate channel-adaptive weights $W_c$ for reweighting. The reweighted feature is denoted as $F_{\text{v}}$. Finally, we aggregate $F_{\text{v}}$, the upsampled higher-scale feature, and the original $F_i^{m}$ via element-wise addition to produce the output feature $F_i$:
\begin{equation}
\left\{
\begin{aligned}
&F_{\text{ref}} = CBR_{3\times 3}\!\big(CBR_{3\times 3}(F_i^{m})\big),\\
&W_c = CA(F_{ref}),
F_{v} = F_{ref} \otimes W_c,\\
&F_i = F_{v} + \mathcal{U}(F_{i+1}, 2) + F_i^{m}.
\end{aligned}
\right.
\tag{9}
\end{equation}

\section{EXPERIMENTS}
\subsection{Experimental Setup}

\noindent
\textbf{Datasets and Evaluation Metrics.}
We evaluate our model on four widely used public COD benchmarks, including CAMO \cite{le2019anabranch}, CHAMELEON \cite{skurowski2018animal}, COD10K \cite{fan2020camouflaged}, and NC4K \cite{lv2021simultaneously}.
Following standard splits \cite{sun2022boundary,pang2022zoom}, we train on 3,040 images from COD10K and 1,000 images from CAMO, and evaluate on the remaining images.
We evaluate performance using four standard metrics: S-measure$(S_{\alpha})$~\cite{fan2017structure}, mean absolute error (MAE), mean E-measure$(E_{\varphi})$~\cite{fan2018enhanced}, and weighted F-measure$(F_\beta^\omega)$~\cite{margolin2014evaluate}.


\begin{figure}[!t]
\centering
\includegraphics[width=\linewidth]{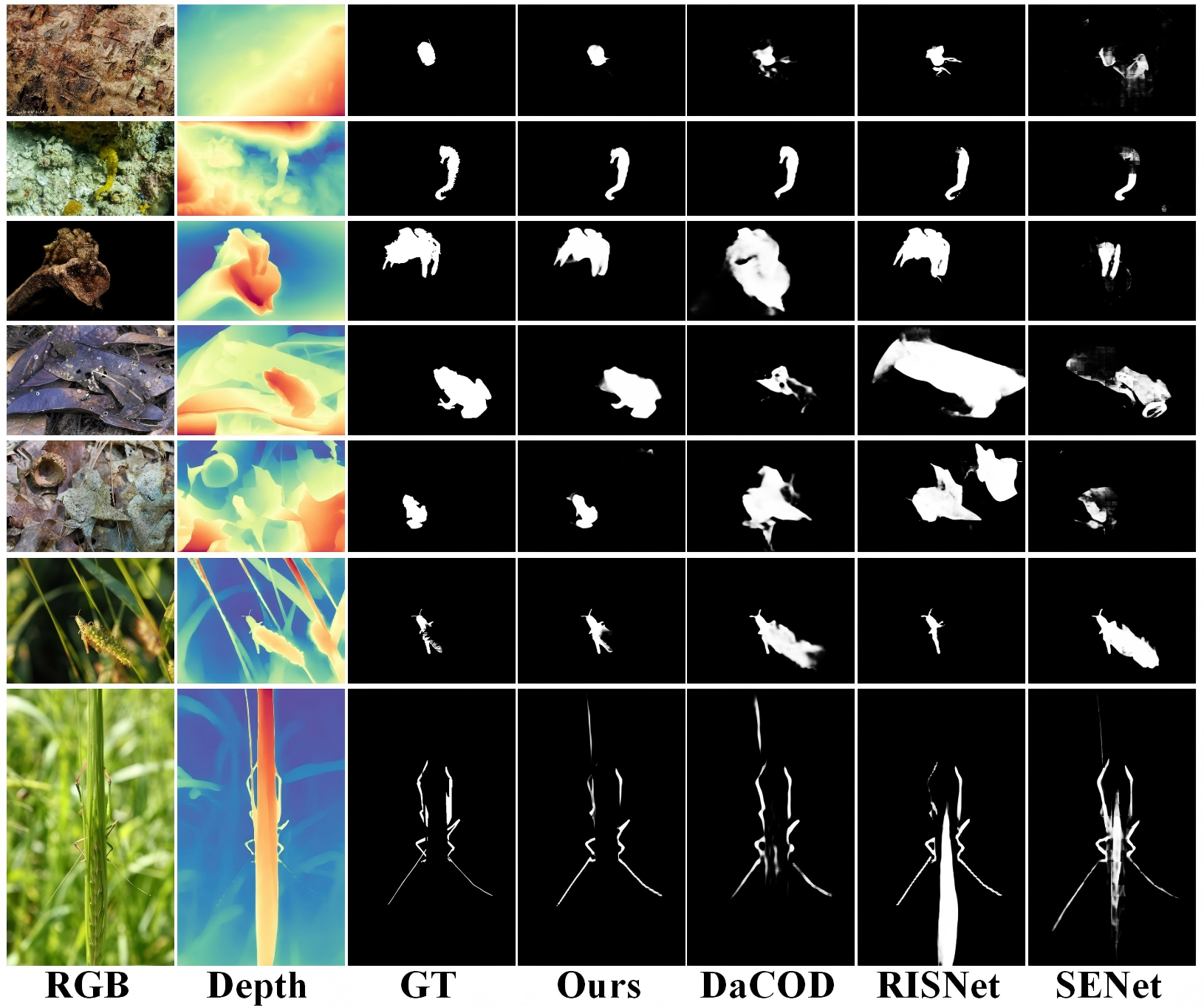}
\vspace{-2mm}
\caption{Visual comparisons of some recent COD methods and ours on different types of samples. More comparisons are provided in the  supplementary material. Best viewed by zooming in for more details.}
\label{fig:visual_comparison}
\end{figure}

\begin{table}[t] 
    \centering
    \setlength{\tabcolsep}{4pt} 
    \caption{Ablation on the THEM, GHEM, and ADFM. B. denotes baseline.} 
    \vspace{-2mm}
    \resizebox{\linewidth}{!}{
    \begin{tabular}{lcccccccccccc} 
        \toprule 
        \multirow{2}{*}{NO.} & \multicolumn{4}{c}{Modules} & \multicolumn{4}{c}{COD10K-Test} & \multicolumn{4}{c}{NC4K-Test} \\
         \cmidrule(lr){2-5}  \cmidrule(lr){6-9}  \cmidrule(lr){10-13} 
        & B. & THEM & GHEM & ADFM & $S_\alpha \uparrow$ & $E_\varphi \uparrow$ & $F_\beta^\omega \uparrow$ & $M \downarrow$ & $S_\alpha \uparrow$ & $E_\varphi \uparrow$ & $F_\beta^\omega \uparrow$ & $M \downarrow$ \\
        \midrule
        1 & $\checkmark$ &  &  &  & 0.873 & 0.925 & 0.786 & 0.022 & 0.893 & 0.930 & 0.842 & 0.032 \\
        2 & $\checkmark$ & $\checkmark$ &  &  & 0.883 & 0.934 & 0.806 & 0.020 & 0.895 & 0.932 & 0.849 & 0.031 \\
        3 & $\checkmark$ &  &  & $\checkmark$ & 0.885 & 0.938 & 0.813 & 0.020 & 0.898 & 0.937 & 0.856 & 0.030 \\
        4 & $\checkmark$ & $\checkmark$ & $\checkmark$ &  & 0.887 & 0.938 & 0.813 & 0.020 & 0.900 & 0.937 & 0.855 & 0.030 \\
        5 & $\checkmark$ & $\checkmark$ & $\checkmark$ & $\checkmark$ & \textcolor{red}{0.889} & \textcolor{red}{0.942} & \textcolor{red}{0.817} & \textcolor{red}{0.019} & \textcolor{red}{0.902} & \textcolor{red}{0.939} & \textcolor{red}{0.859} & \textcolor{red}{0.029} \\
        \bottomrule 
    \end{tabular}
    }
\label{tab:ablation}
\vspace{-2mm}
\end{table}

\noindent
\textbf{Implementation Details.} 
In the training and testing stages, we resize the RGB and depth inputs to $416\times416$. During training, we employ data augmentation strategies such as random flipping, rotation, and cropping to prevent overfitting. We train with Adam~\cite{adam2014method} for a total of 100 epochs, using a batch size of 8 and an initial learning rate of $5\times10^{-5}$, and we decay the learning rate by a factor of 10 every 40 epochs. Our model is trained on two RTX2080 GPUs. 

\noindent
\subsection{Comparison with SOTAs.}
\noindent
\textbf{Quantitative Evaluation.}
To evaluate MHENet, we compare it with 16 SOTAs on four public benchmark datasets.
These methods include SINet \cite{fan2020camouflaged}, BGNet \cite{sun2022boundary}, ZoomNet \cite{pang2022zoom}, FSPNet \cite{huang2023feature}, UEDG \cite{lyu2023uedg}, MSCAF \cite{liu2023mscaf}, DINet \cite{zhou2024decoupling}, RISNet \cite{wang2024depth}, ICEG \cite{he2023strategic}, DSNet \cite{lyu2025distraction}, PRBENet \cite{yue2024progressive}, SENet \cite{hao2025simple}, DaCOD \cite{wang2023depth}, DSAM \cite{yu2024exploring}, MAGNet \cite{zhong2024magnet}, MultiCOS \cite{fang2025integrating}. 
All results are taken from published papers or reproduced using public code under our setups for fair comparison.
As shown in Tab.~\ref{tab:comparison}, our MHENet achieves the best or highly competitive overall performance across all benchmarks.

\begin{table}[t]
    \centering
    \setlength{\tabcolsep}{4pt}
    \caption{Ablation Study of the Key Block in THEM and GHEM.}
    \vspace{-2mm}
    \resizebox{\linewidth}{!}{
    \begin{tabular}{llcccccccc}
        \toprule
        \multirow{2}{*}{NO.} & 
        \multirow{2}{*}{Inputs} & \multicolumn{4}{c}{COD10K-Test} & \multicolumn{4}{c}{NC4K-Test} \\
         \cmidrule(lr){3-6} \cmidrule(lr){7-10}
        & & $S_\alpha \uparrow$ & $E_\varphi \uparrow$ & $F_\beta^\omega \uparrow$ & $M \downarrow$ & $S_\alpha \uparrow$ & $E_\varphi \uparrow$ & $F_\beta^\omega \uparrow$ & $M \downarrow$ \\
        \midrule
        6 & w/o Semantic Block & 0.885 & 0.939 & 0.816 & 0.020 & 0.898 & 0.935 & 0.856 & 0.030 \\
        7 & w/o Texture Block & 0.886 & 0.940 & 0.814 & 0.020 & 0.899 & 0.937 & 0.855 & 0.030 \\
        8 & w/o Geometry Block & 0.887 & 0.941 & 0.815 & 0.020 & 0.899 & 0.937 & 0.856 & 0.030 \\
        9 & MHENet (Ours) & \textcolor{red}{0.889} & \textcolor{red}{0.942} & \textcolor{red}{0.817} & \textcolor{red}{0.019} &\textcolor{red}{0.902} & \textcolor{red}{0.939} & \textcolor{red}{0.859} & \textcolor{red}{0.029} \\
            \bottomrule
        \end{tabular}
        }
\label{tab:ablationOfComponent}
\end{table}

\begin{table}[t]
    \centering
    \setlength{\tabcolsep}{4pt}
    \caption{Ablation Study of the Inputs of MHENet.}
    \vspace{-2mm}
    \resizebox{\linewidth}{!}{
    \begin{tabular}{llcccccccc}
        \toprule
        \multirow{2}{*}{NO.} & 
        \multirow{2}{*}{Inputs} & \multicolumn{4}{c}{COD10K-Test} & \multicolumn{4}{c}{NC4K-Test} \\
         \cmidrule(lr){3-6} \cmidrule(lr){7-10}
        & & $S_\alpha \uparrow$ & $E_\varphi \uparrow$ & $F_\beta^\omega \uparrow$ & $M \downarrow$ & $S_\alpha \uparrow$ & $E_\varphi \uparrow$ & $F_\beta^\omega \uparrow$ & $M \downarrow$ \\
        \midrule
        10 & only Depth & 0.844 & 0.902 & 0.740 & 0.028 & 0.869 & 0.912 & 0.807 & 0.039 \\
        11 & only RGB & 0.880 & 0.937 & 0.803 & 0.021 & 0.895 & 0.936 & 0.851 & 0.029 \\
    12 & RGB-D & \textcolor{red}{0.889} & \textcolor{red}{0.942} & \textcolor{red}{0.817} & \textcolor{red}{0.019} &\textcolor{red}{0.902} & \textcolor{red}{0.939} & \textcolor{red}{0.859} & \textcolor{red}{0.029} \\
        \bottomrule
    \end{tabular}
    }
\label{tab:ablationOfInputs}
\vspace{-2mm}
\end{table}

\noindent
\textbf{Qualitative Evaluation.}
Visual comparisons of different methods on several representative samples are shown in Fig.~\ref{fig:visual_comparison}. These cases highlight diverse challenges, including ambiguous depth (Row 1), small objects (Row 2), indistinct boundaries (Row 3), background interference (Rows 4-5), and occlusion (Rows 6-7). Overall, our method produces more accurate segmentation results across these challenging scenarios.

\subsection{Ablation Study}

\noindent
\textbf{Effectiveness of THEM, GHEM, and ADFM.}
From Tab.~\ref{tab:ablation}, THEM (NO. 2) improves the baseline (NO. 1) with $F_\beta^\omega$ gains (2.5\%, on COD10K-Test). Adding GHEM (NO. 4) brings further consistent improvements, and ADFM (NO. 3, 5) outperforms convolution-based fusion and achieves the best overall results.

\noindent
\textbf{Effectiveness of Semantic, Texture, and Geometry Blocks.} 
From Tab.~\ref{tab:ablationOfComponent}, removing any block consistently degrades performance on COD10K-Test and NC4K-Test, while MHENet achieves the best results.
The Semantic, Texture, and Geometry Blocks respectively maintain cross-scale consistency, refine RGB textures, and reinforce depth structures, complementing each other in RGB-D COD.

\noindent
\textbf{Effectiveness of Different Inputs.}
From Tab.~\ref{tab:ablationOfInputs}, the input modality has an impact on performance. RGB outperforms depth, while RGB-D fusion achieves the best results by leveraging complementary texture and geometric cues.

\section{CONCLUSION}
In this paper, we propose MHENet, a modality-specific hierarchical enhancement framework for RGB-D camouflaged object detection. 
MHENet strengthens RGB texture cues and depth geometric cues with THEM and GHEM, while a semantic block maintains cross-scale semantic consistency for both RGB features and depth features.
Then, ADFM selectively fuses the enhanced features via spatially adaptive weighting, enabling the network to emphasize the more reliable modality. 
Extensive experiments on four benchmarks demonstrate that MHENet outperforms 16 state-of-the-art methods.


\section*{Acknowledgments}
This work was supported in part by the National Natural Science Foundation of China under Grant 62471142, in part by the Natural Science Foundation of Fujian Province, China under Grant 2023J01067, in part by the Industry-Academy Cooperation Project under Grant 2024H6006, in part by the Collaborative Innovation Platform Project of Fuzhou City under Grant 2023-P-002, in part by the Key Technology Innovation Project for Focused Research and Industrialization in the Software Industry of Fujian Province, and in part by the Fuzhou University Startup Funding 511704.

\putbib
\end{bibunit}

\clearpage
\appendices
\renewcommand{\thesection}{S\arabic{section}}
\renewcommand{\thefigure}{S\arabic{figure}}
\renewcommand{\thetable}{S\arabic{table}}
\setcounter{section}{0}
\setcounter{figure}{0}
\setcounter{table}{0}
\section*{Supplementary Material}
\addcontentsline{toc}{section}{Supplementary Material}

\begin{bibunit}


In this supplementary material, we provide the following materials:
\begin{itemize}
    \item Related work on COD and RGB-D COD
    \item Experimental setup details, including the loss function, evaluation metrics and datasets;
    \item Robustness to different depth estimation models;
    \item Complexity analysis;
    \item Limitations and future work;
    \item More feature visualizations;
    \item Additional comparisons.

\end{itemize}

\section{RELATED WORK}

\subsection{Camouflaged Object Detection}
Early COD methods largely relied on hand-crafted features, such as intensity~\cite{tankus2001convexity}, color~\cite{huerta2007improving} and texture~\cite{sengottuvelan2008performance}. In contrast, more recent deep learning methods automatically learn representations that capture complex features directly from data, 
demonstrating superior performance even in complex image segmentation tasks \cite{zeng2024mgqformer,zhang2025excitation}. This progress has been greatly facilitated by the establishment of large-scale datasets. Pioneer efforts introduced several key benchmarks, including CAMO~\cite{le2019anabranch}, COD10K~\cite{fan2020camouflaged}, and NC4K~\cite{lv2021simultaneously}. These datasets have become the standard benchmarks for the COD field, providing a solid foundation for training and evaluating a wide range of subsequent deep learning models. 

Building upon these foundational datasets, a variety of deep learning-based COD frameworks have been proposed to better capture the subtle and concealed visual cues that are characteristic of camouflaged scenes. Sun \textit{et al.}~\cite{sun2022boundary} proposed a boundary-guided network (BGNet) that leverages object-related edge semantics to guide the representation learning of camouflaged objects, effectively addressing the problem of incomplete boundaries in COD. Pang \textit{et al.}~\cite{pang2022zoom} proposed a mixed-scale triplet network (ZoomNet), which mimics human behavior of zooming in and out to address challenges like diverse object scales, high foreground-background similarity, and ambiguous predictions. Recently, SAM-based frameworks~\cite{Ren_2025_ICCV},~\cite{gao2025cod} have attracted increasing attention. Several works adapt SAM to camouflaged object detection through prompt refinement and structural priors. These adaptations demonstrate SAM's strong potential for improving boundary localization and structural consistency.

\subsection{RGB-D Camouflaged Object Detection}
In RGB-D camouflaged object detection, RGB images provide rich color and texture information, whereas depth images emphasize three-dimensional layout and spatial positional information. A central challenge in this field remains how to effectively integrate these complementary RGB and depth features to achieve robust cross-modal fusion. 

Numerous studies have been devoted to addressing this issue. Wang \textit{et al.}~\cite{wang2023depth} proposed a Depth-aided Camouflaged Object Detection Framework (DaCOD) that incorporates depth information as a complementary cue, enabling multi-modal collaborative learning through a hybrid backbone (SwinL and ResNet50) and adopting an asymmetric cross-modal fusion strategy to transfer informative RGB features to the depth branch for enhanced feature representation. Liu \textit{et al.}~\cite{liu2024depth} proposed a Depth-perception Attention Fusion Network (DAF-Net), which adpots a three-branch encoder, a depth-weighted cross-attention fusion module to adjust fusion weights dynamically, and a feature aggregation decoder to fuse enhanced features for accurate segmentation. Wang \textit{et al.}~\cite{wang2024depth} proposed a Recurrent Iterative Segmentation Network (RISNet), which integrates multi-scale RGB features with depth-guided spatial information through iterative refinement to enhance concealed crop detection in dense agricultural scenes.

Different from these methods that mainly focus on designing sophisticated cross-modal attention or iterative fusion strategies, our approach explicitly considers the intrinsic characteristics of RGB and depth modalities, and performs modality-specific feature modeling followed by a more targeted and adaptive cross-modal fusion.

\section{Experimental Setup}
\subsection{Loss Function}
In this paper, we apply supervision on the prediction masks $M_1$, $M_2$ and $M_3$
 from the features $R_1$, $F_1$ and $D_1$
 using a hybrid loss that consists of BCE loss \cite{de2005tutorial} and IoU loss \cite{mattyus2017deeproadmapper}.
The hybrid loss supervises predictions at the pixel level as well as the foreground and background regions. 
BCE loss is defined as:
\begin{equation}
\begin{split}
\mathcal{L}_{BCE} &= -\sum_{x=1}^{H} \sum_{y=1}^{W} \Big[ G(x,y)\log(M_i(x,y)) \\
&\quad + \big(1 - G(x,y)\big)\log\big(1 - M_i(x,y)\big) \Big],
\end{split}
\end{equation}
where $W$ and $H$ represent the width and height of the image, respectively, $M_i$ is the predicted mask, and $G$ is the ground truth mask. 

IoU loss is defined as:
{\small
\begin{equation}
\mathcal{L}_{IoU} = 1 - \frac{\sum_{x=1}^{H} \sum_{y=1}^{W} M_i(x,y)G(x,y)}{\sum_{x=1}^{H} \sum_{y=1}^{W} \Big[ M_i(x,y) + G(x,y) - M_i(x,y)G(x,y) \Big]}
\end{equation}
}

The overall loss $\mathcal{L}$ of the model is defined as:
\begin{equation}
\begin{split}
\mathcal{L} &= \sum_{i=1}^{3} \left[ \mathcal{L}_{BCE}^i (M_i, G) + \mathcal{L}_{IoU}^i (M_i, G) \right].
\end{split}
\end{equation}

\begin{figure}[t]
\centering
\includegraphics[width=\linewidth]{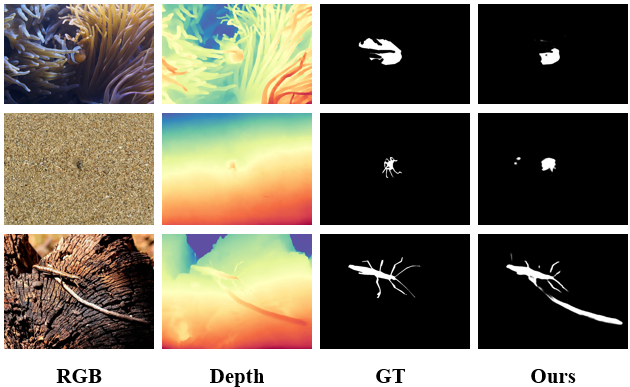}
\caption{Failure cases and potential extensions of MHENet under occlusion, ambiguous boundaries, and noisy depth scenarios.}
\label{fig:limitations}
\end{figure}

\subsection{Evaluation Metrics}
We evaluate the prediction performance using four widely adopted metrics, including mean absolute error (MAE), weighted F-measure ($F_{\beta}^{\omega}$)~\cite{margolin2014evaluate}, mean E-measure ($E_{\varphi}$)~\cite{fan2018enhanced}, and S-measure ($S_{\alpha}$)~\cite{fan2017structure}. Among them, MAE evaluates the average pixel-wise prediction error, where a lower value indicates better performance. In contrast, $F_{\beta}^{\omega}$, $E_{\varphi}$, and $S_{\alpha}$ measure the similarity between the prediction mask and the ground-truth mask, and higher values indicate better performance.

Let $M \in [0,1]^{W \times H}$ denote the prediction mask and $G \in \{0,1\}^{W \times H}$ denote the binary ground-truth mask, where $W$ and $H$ represent the width and height of the image, respectively.

MAE measures the average absolute difference between the prediction mask and the ground-truth mask:
\begin{equation}
\mathrm{MAE} = \frac{1}{W \times H}\sum_{x=1}^{H}\sum_{y=1}^{W} \left| M(x,y)-G(x,y) \right|,
\end{equation}
where $x$ and $y$ denote the pixel indices along the height and width dimensions, respectively. The value of MAE is normalized to $[0,1]$, and a smaller value indicates better prediction quality.

Weighted F-measure is an improved variant of the traditional F-measure, which introduces spatial weighting to better reflect the perceptual importance of different prediction errors~\cite{margolin2014evaluate}. It is defined as
\begin{equation}
F_{\beta}^{\omega} = \frac{(1+\beta^2)\mathrm{Precision}^{\omega}\times \mathrm{Recall}^{\omega}}
{\beta^2 \mathrm{Precision}^{\omega}+\mathrm{Recall}^{\omega}},
\end{equation}
where $\mathrm{Precision}^{\omega}$ and $\mathrm{Recall}^{\omega}$ denote the weighted precision and weighted recall.

Mean E-measure jointly considers the global image-level statistics and local pixel-level matching information~\cite{fan2018enhanced}. It is defined as
\begin{equation}
E_{\varphi} = \frac{1}{W \times H}\sum_{x=1}^{H}\sum_{y=1}^{W}\varphi(x,y),
\end{equation}
where $\varphi(x,y)$ denotes the enhanced alignment value at pixel $(x,y)$, reflecting the alignment consistency between the prediction mask and the ground-truth mask. A larger $E_{\varphi}$ value indicates better consistency between $M$ and $G$.

S-measure evaluates the structural similarity between the prediction mask and the ground-truth mask by jointly considering object-aware and region-aware structural information~\cite{fan2017structure}. It is defined as
\begin{equation}
S_{\alpha} = \alpha S_o + (1-\alpha)S_r,
\end{equation}
where $S_o$ denotes the object-aware structural similarity, $S_r$ denotes the region-aware structural similarity, and $\alpha$ is a balancing factor. 

\subsection{Datasets}
We evaluate our model on four widely used public camouflaged object detection (COD) benchmarks: CAMO \cite{le2019anabranch}, CHAMELEON \cite{skurowski2018animal}, COD10K\cite{fan2020camouflaged}, and NC4K\cite{lv2021simultaneously}.  
CAMO contains 1,250 camouflaged images (1,000 for training and 250 for testing), covering both natural and artificial camouflaged objects. 
CHAMELEON is a small-scale test-only dataset with 76 images, focusing on animals camouflaged in complex ecological backgrounds. 
As a comprehensive large-scale dataset, COD10K includes 5,066 camouflaged images (3,040 for training and 2,026 for testing), along with 1,934 non-camouflaged and 3,000 background images, supported by high-quality annotations for multiple COD-related tasks. 
NC4K, the largest existing COD test dataset, consists of 4,121 camouflaged scene images with object and instance-level annotations, suitable for evaluating model generalization. 
Following the standard training-test splits adopted in previous works, 3040 images from COD10K and 1000 images from CAMO are used for training, and the rest ones are used for testing.

\begin{figure}[t]
\centering
\includegraphics[width=\linewidth]
{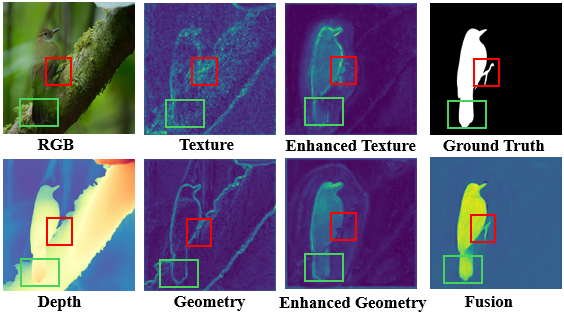}
\includegraphics[width=\linewidth]
{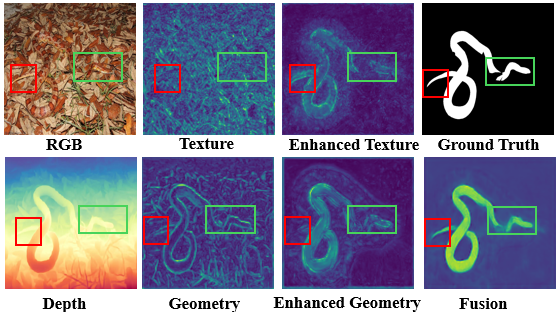}
\includegraphics[width=\linewidth]
{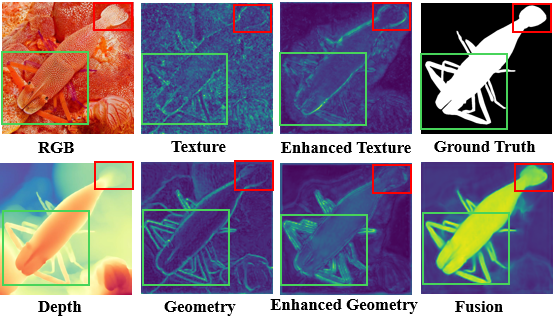}
\caption{Visualization of the features of our MHENet. Best viewed by zooming in for more details.}
\label{fig:features}
\end{figure}

\begin{figure*}[t]
\centering
\includegraphics[width=\linewidth]
{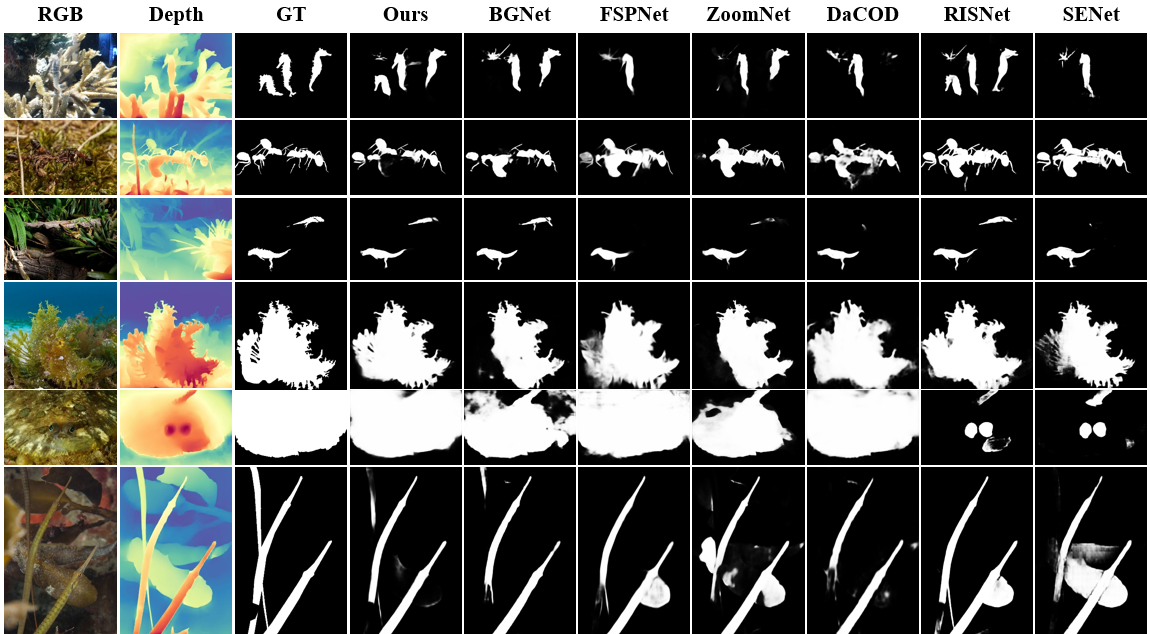}
\includegraphics[width=\linewidth]
{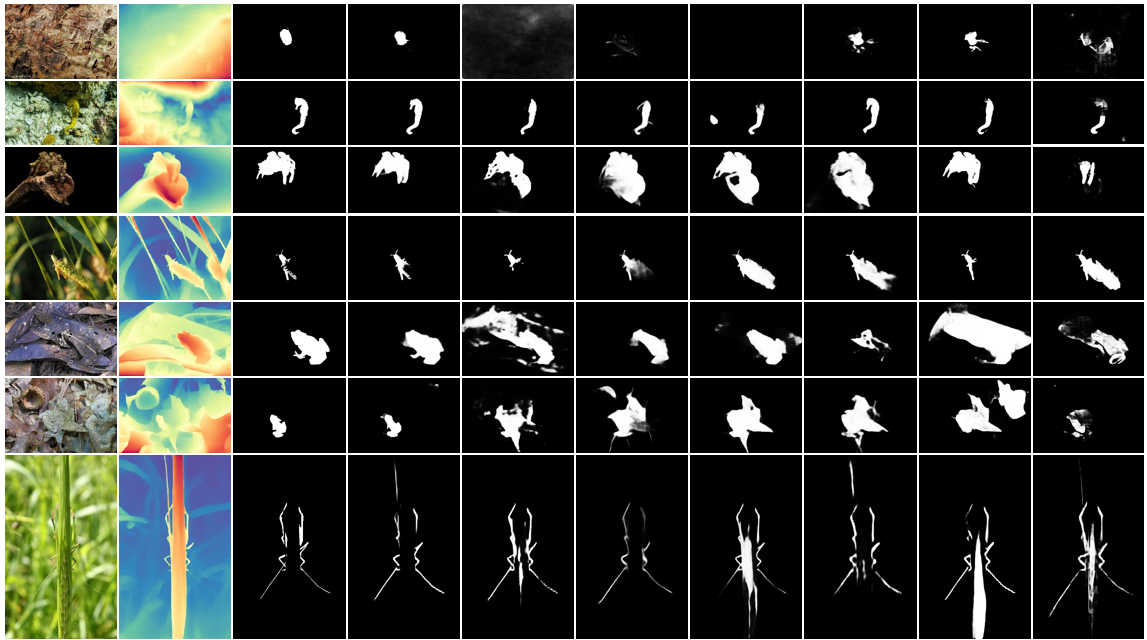}
\includegraphics[width=\linewidth]
{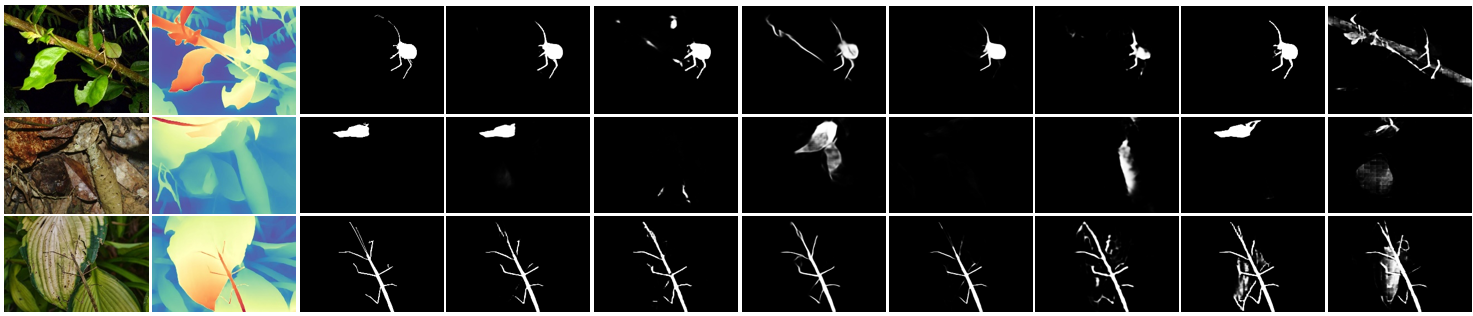}
\caption{Visual comparisons of some recent COD methods and ours on different types of samples. Please zoom in for more details.}
\label{fig:comparison}
\end{figure*}

\section{Robustness to different depth estimation models.}
In the main paper, the depth maps are generated by Distill Any Depth \cite{he2025distill}. To study the sensitivity of our method to the choice of depth estimator, we further replace it with Depth Anything V1 \cite{yang2024depth} and Depth Anything V2 \cite{yang2024depth2}. The quantitative results are presented in Tab.~\ref{tab:DepthEst}. It can be observed that the performance remains largely consistent across different depth estimation models on both COD10K-Test and NC4K-Test, which demonstrates the robustness of our method.

\begin{table}[t]
\centering
\setlength{\tabcolsep}{4pt}
\caption{Performance comparison using different depth estimation models.}
\resizebox{\linewidth}{!}{
\begin{tabular}{lcccccccc}
\toprule
\multirow{2}{*}{Depth Estimation Model} & \multicolumn{4}{c}{COD10K-Test} & \multicolumn{4}{c}{NC4K-Test} \\
\cmidrule(lr){2-5} \cmidrule(lr){6-9}
& $S_\alpha \uparrow$ & $E_\varphi \uparrow$ & $F_\beta^\omega \uparrow$ & $M \downarrow$
& $S_\alpha \uparrow$ & $E_\varphi \uparrow$ & $F_\beta^\omega \uparrow$ & $M \downarrow$ \\
\midrule
Depth Anything V1  & 0.886 & 0.939 & 0.812 & 0.020 & 0.900 & 0.937 & 0.857 & 0.030 \\
Depth Anything V2  & 0.894 & 0.941 & 0.825 & 0.019 & 0.905 & 0.941 & 0.865 & 0.029 \\
Distill Any Depth  & 0.889 & 0.942 & 0.817 & 0.019 & 0.902 & 0.939 & 0.859 & 0.029 \\
\bottomrule
\end{tabular}
}
\label{tab:DepthEst}
\vspace{-2mm}
\end{table}

\begin{table}[t]
\centering
\caption{Complexity analysis of different methods.}
\resizebox{0.40\textwidth}{!}{
\begin{tabular}{lccccc}
\toprule
Methods/Metrics & SINet & BGNet & FSPNet & RISNet & Ours \\
\midrule
Params (M) $\downarrow$ & 48.95 & 79.85 & 273.80 & 26.58 & 53.36 \\
FLOPs (G) $\downarrow$  & 38.76 & 59.45 & 283.31 & 96.57 & 37.97 \\
Speed (FPS) $\uparrow$  & 46.06 & 39.49 & 26.23 & 24.82 & 23.43 \\
\bottomrule
\end{tabular}
}
\label{tab:Complexity}
\vspace{-7mm}
\end{table}

\section{Complexity analysis.}
We further compare the model complexity of our method with several representative methods in terms of parameter count, FLOPs, and inference speed. As reported in Tab.~\ref{tab:Complexity}, our method maintains a moderate number of parameters and competitive inference speed, while achieving the lowest FLOPs among the compared methods. These results indicate that our method provides a favorable trade-off between computational cost and practical efficiency.

\section{Limitations and Future Work}
As shown in Fig.~\ref{fig:limitations}, MHENet still faces difficulties in several challenging cases. When the camouflaged object is heavily occluded by surrounding background (e.g., the small creature hidden in the anemone in Row~1), the predicted mask may be less complete. A promising direction is to incorporate a background-aware modeling strategy that better characterizes background statistics and improves target–background separation under occlusion. For targets with ambiguous boundaries (e.g., the tiny insect blending into the sandy background in Row~2), the prediction can exhibit slightly blurred edges or minor redundant regions, which could be further alleviated by introducing lightweight edge-aware supervision to sharpen boundary representations. In addition, when the depth map contains artifacts (e.g., distorted depth cues around the lizard in Row~3), the contour may deviate locally. This suggests exploring reliability-aware depth refinement, such as an attention gating mechanism, to down-weight noisy depth features and enhance robustness.

\section{More Feature Visualizations}
We provide additional feature visualizations (Fig. \ref{fig:features}) in this supplementary material to further demonstrate the effectiveness of our proposed modality-specific hierarchical enhancement modules.

\section{Additional Comparisons}
Due to space limitations of the manuscript, we add more
visual comparisons (see Fig. \ref{fig:comparison}) to this supplementary material to further demonstrate the performance of our model.
The methods used in the experiments for visual comparison include 
BGNet\cite{sun2022boundary}, FSPNet \cite{huang2023feature}, ZoomNet \cite{pang2022zoom}, DaCOD \cite{wang2023depth},  RISNet \cite{wang2024depth}, SENet \cite{hao2025simple}.

Our method delivers strong performance across a range of diverse scenarios: for scenes with multiple objects (Rows 1-3), it accurately captures all target instances unlike some competitors that miss certain objects; for large-sized objects (Rows 4-5), it preserves the full shape and contour of targets; for elongated objects (Rows 6 and 16), it retains the slender structures of targets while competing approaches often show structural distortion or incomplete segmentation; in cases with ambiguous depth (Row 7), it effectively distinguishes targets from the background by integrating depth and texture information as opposed to others’ confusing segmentation results; for small-sized objects (Rows 8, 14, and 15), it precisely locates and segments these tiny targets; for scenes with blurry boundaries (Row 9), it defines clear target edges; for occluded objects (Rows 10 and 13), it successfully identifies intact targets despite partial occlusion; and in cluttered background scenes (Rows 11-12), it reliably separates targets from complex background noise while other approaches make errors due to background distraction.


\putbib
\end{bibunit}


\begin{thebibliography}{10}

\bibitem{price2019background}
Natasha Price, Samuel Green, Jolyon Troscianko, Tom Tregenza, and Martin Stevens,
\newblock ``Background matching and disruptive coloration as habitat-specific strategies for camouflage,''
\newblock vol. 9, pp. 7840, 2019.

\bibitem{stevens2009animal}
Martin Stevens and Sami Merilaita,
\newblock ``Animal camouflage: Current issues and new perspectives,''
\newblock vol. 364, pp. 423--427, 2009.

\bibitem{fan2020camouflaged}
Deng-Ping Fan, Ge-Peng Ji, Guolei Sun, Ming-Ming Cheng, Jianbing Shen, and Ling Shao,
\newblock ``Camouflaged object detection,''
\newblock in {\em Proc. IEEE/CVF Conf. Comput. Vis. Pattern Recognit.}, 2020, pp. 2777--2787.

\bibitem{cui2021sddnet}
Lisha Cui, Xiaoheng Jiang, Mingliang Xu, Wanqing Li, Pei Lv, and Bing Zhou,
\newblock ``{SDDNet}: A fast and accurate network for surface defect detection,''
\newblock {\em IEEE Trans. Instrum. Meas.}, vol. 70, pp. 1--13, 2021.

\bibitem{rustia2020application}
Dan Jeric~Arcega Rustia, Chien~Erh Lin, Jui-Yung Chung, Yi-Ji Zhuang, Ju-Chun Hsu, and Ta-Te Lin,
\newblock ``Application of an image and environmental sensor network for automated greenhouse insect pest monitoring,''
\newblock vol. 23, pp. 17--28, 2020.

\bibitem{fan2020pranet}
Deng-Ping Fan, Ge-Peng Ji, Tao Zhou, Geng Chen, Huazhu Fu, Jianbing Shen, and Ling Shao,
\newblock ``{PraNet}: Parallel reverse attention network for polyp segmentation,''
\newblock in {\em Proc. Int. Conf. Med. Image Comput. Comput.-Assist. Interv.}, 2020, pp. 263--273.

\bibitem{perez2012early}
Ricardo P{\'e}rez-de~la Fuente, Xavier Delclos, Enrique Pe{\~n}alver, Mariela Speranza, Jacek Wierzchos, Carmen Ascaso, and Michael~S Engel,
\newblock ``Early evolution and ecology of camouflage in insects,''
\newblock vol. 109, pp. 21414--21419, 2012.

\bibitem{sun2022boundary}
Yujia Sun, Shuo Wang, Chenglizhao Chen, and Tian-Zhu Xiang,
\newblock ``Boundary-guided camouflaged object detection,''
\newblock in {\em Proc. Int. Joint Conf. Artif. Intell.}, 2022, pp. 1335--1341.

\bibitem{yue2024progressive}
Guanghui Yue, Shangjie Wu, Tianwei Zhou, Gang Li, Jie Du, Yu~Luo, and Qiuping Jiang,
\newblock ``Progressive region-to-boundary exploration network for camouflaged object detection,''
\newblock {\em IEEE Trans. Multimedia}, vol. 27, pp. 236--248, 2024.

\bibitem{pang2022zoom}
Youwei Pang, Xiaoqi Zhao, Tian-Zhu Xiang, Lihe Zhang, and Huchuan Lu,
\newblock ``Zoom in and out: A mixed-scale triplet network for camouflaged object detection,''
\newblock in {\em Proc. IEEE/CVF Conf. Comput. Vis. Pattern Recognit.}, 2022, pp. 2160--2170.

\bibitem{lyu2025distraction}
Han Lyu, Meijun Sun, Haowei Ran, Yipu Liu, Xinyu Yan, and Zheng Wang,
\newblock ``Distraction suppression and feature modulation network for camouflaged object detection,''
\newblock in {\em Proc. IEEE Int. Conf. Multimedia Expo}, 2025, pp. 1--6.

\bibitem{wang2023depth}
Qingwei Wang, Jinyu Yang, Xiaosheng Yu, Fangyi Wang, Peng Chen, and Feng Zheng,
\newblock ``Depth-aided camouflaged object detection,''
\newblock in {\em Proc. ACM Int. Conf. Multimedia}, 2023, pp. 3297--3306.

\bibitem{wang2024depth}
Liqiong Wang, Jinyu Yang, Yanfu Zhang, Fangyi Wang, and Feng Zheng,
\newblock ``Depth-aware concealed crop detection in dense agricultural scenes,''
\newblock in {\em Proc. IEEE/CVF Conf. Comput. Vis. Pattern Recognit.}, 2024, pp. 17201--17211.

\bibitem{hu2024cross}
Xihang Hu, Fuming Sun, Jing Sun, Fasheng Wang, and Haojie Li,
\newblock ``Cross-modal fusion and progressive decoding network for rgb-d salient object detection,''
\newblock {\em Int. J. Comput. Vis.}, vol. 132, pp. 3067--3085, 2024.

\bibitem{galun2003texture}
Galun, Sharon, Basri, and Brandt,
\newblock ``Texture segmentation by multiscale aggregation of filter responses and shape elements,''
\newblock in {\em Proc. IEEE/CVF Int. Conf. Comput. Vis.}, 2003, pp. 716--723.

\bibitem{wu2023source}
Zongwei Wu, Danda~Pani Paudel, Deng-Ping Fan, Jingjing Wang, Shuo Wang, C{\'e}dric Demonceaux, Radu Timofte, and Luc Van~Gool,
\newblock ``Source-free depth for object pop-out,''
\newblock in {\em Proc. IEEE/CVF Int. Conf. Comput. Vis.}, 2023, pp. 1032--1042.

\bibitem{he2025distill}
Xiankang He, Dongyan Guo, Hongji Li, Ruibo Li, Ying Cui, and Chi Zhang,
\newblock ``Distill any depth: Distillation creates a stronger monocular depth estimator,''
\newblock {\em arXiv preprint arXiv: 2502.19204}, 2025.

\bibitem{wang2021pvtv2}
Wenhai Wang, Enze Xie, Xiang Li, Deng-Ping Fan, Kaitao Song, Ding Liang, Tong Lu, Ping Luo, and Ling Shao,
\newblock ``{PVTv2}: Improved baselines with pyramid vision transformer,''
\newblock {\em Comput. Visual Media}, vol. 8, pp. 1--10, 2022.

\bibitem{huang2023feature}
Zhou Huang, Hang Dai, Tian-Zhu Xiang, Shuo Wang, Huai-Xin Chen, Jie Qin, and Huan Xiong,
\newblock ``Feature shrinkage pyramid for camouflaged object detection with transformers,''
\newblock in {\em Proc. IEEE/CVF Conf. Comput. Vis. Pattern Recognit.}, 2023, pp. 5557--5566.

\bibitem{lyu2023uedg}
Yixuan Lyu, Hong Zhang, Yan Li, Hanyang Liu, Yifan Yang, and Ding Yuan,
\newblock ``{UEDG}: Uncertainty-edge dual guided camouflage object detection,''
\newblock {\em IEEE Trans. Multimedia}, vol. 26, pp. 4050--4060, 2023.

\bibitem{liu2023mscaf}
Yu~Liu, Haihang Li, Juan Cheng, and Xun Chen,
\newblock ``{MSCAF-Net}: A general framework for camouflaged object detection via learning multi-scale context-aware features,''
\newblock {\em IEEE Trans. Circuits Syst. Video Technol.}, vol. 33, pp. 4934--4947, 2023.

\bibitem{zhou2024decoupling}
Xiaofei Zhou, Zhicong Wu, and Runmin Cong,
\newblock ``Decoupling and integration network for camouflaged object detection,''
\newblock {\em IEEE Trans. Multimedia}, vol. 26, pp. 7114--7129, 2024.

\bibitem{he2023strategic}
Chunming He, Kai Li, Yachao Zhang, Yulun Zhang, Zhenhua Guo, Xiu Li, Martin Danelljan, and Fisher Yu,
\newblock ``Strategic preys make acute predators: Enhancing camouflaged object detectors by generating camouflaged objects,''
\newblock {\em arXiv preprint arXiv:2308.03166}, 2023.

\bibitem{hao2025simple}
Chao Hao, Zitong Yu, Xin Liu, Jun Xu, Huanjing Yue, and Jingyu Yang,
\newblock ``A simple yet effective network based on vision transformer for camouflaged object and salient object detection,''
\newblock {\em IEEE Trans. Image Process.}, vol. 34, pp. 608--622, 2025.

\bibitem{yu2024exploring}
Zhenni Yu, Xiaoqin Zhang, Li~Zhao, Yi~Bin, and Guobao Xiao,
\newblock ``Exploring deeper! {Segment} anything model with depth perception for camouflaged object detection,''
\newblock in {\em Proc. ACM Int. Conf. Multimedia}, 2024, pp. 4322--4330.

\bibitem{zhong2024magnet}
Mingyu Zhong, Jing Sun, Peng Ren, Fasheng Wang, and Fuming Sun,
\newblock ``{MAGNet}: Multi-scale awareness and global fusion network for rgb-d salient object detection,''
\newblock {\em Knowl.-Based Syst.}, vol. 299, pp. 112126, 2024.

\bibitem{fang2025integrating}
Chengyu Fang, Chunming He, Longxiang Tang, Yuelin Zhang, Chenyang Zhu, Yuqi Shen, Chubin Chen, Guoxia Xu, and Xiu Li,
\newblock ``Integrating extra modality helps segmentor find camouflaged objects well,''
\newblock {\em arXiv preprint arXiv:2502.14471}, 2025.

\bibitem{le2019anabranch}
Trung-Nghia Le, Tam~V Nguyen, Zhongliang Nie, Minh-Triet Tran, and Akihiro Sugimoto,
\newblock ``Anabranch network for camouflaged object segmentation,''
\newblock {\em Comput. Vis. Image Underst.}, vol. 184, pp. 45--56, 2019.

\bibitem{skurowski2018animal}
Przemys{\l}aw Skurowski, Hassan Abdulameer, Jakub B{\l}aszczyk, Tomasz Depta, Adam Kornacki, and Przemys{\l}aw Kozie{\l},
\newblock ``Animal camouflage analysis: Chameleon database,'' 2018,
\newblock Unpublished Manuscript.

\bibitem{lv2021simultaneously}
Yunqiu Lv, Jing Zhang, Yuchao Dai, Aixuan Li, Bowen Liu, Nick Barnes, and Deng-Ping Fan,
\newblock ``Simultaneously localize, segment and rank the camouflaged objects,''
\newblock in {\em Proc. IEEE/CVF Conf. Comput. Vis. Pattern Recognit.}, 2021, pp. 11591--11601.

\bibitem{fan2017structure}
Deng-Ping Fan, Ming-Ming Cheng, Yun Liu, Tao Li, and Ali Borji,
\newblock ``Structure-measure: A new way to evaluate foreground maps,''
\newblock in {\em Proc. IEEE/CVF Int. Conf. Comput. Vis.}, 2017, pp. 4548--4557.

\bibitem{fan2018enhanced}
Deng-Ping Fan, Cheng Gong, Yang Cao, Bo~Ren, Ming-Ming Cheng, and Ali Borji,
\newblock ``Enhanced-alignment measure for binary foreground map evaluation,''
\newblock {\em arXiv preprint arXiv:1805.10421}, 2018.

\bibitem{margolin2014evaluate}
Ran Margolin, Lihi Zelnik-Manor, and Ayellet Tal,
\newblock ``How to evaluate foreground maps?,''
\newblock in {\em Proc. IEEE/CVF Conf. Comput. Vis. Pattern Recognit.}, 2014, pp. 248--255.

\bibitem{adam2014method}
Diederik Kinga, Jimmy~Ba Adam, et~al.,
\newblock ``A method for stochastic optimization,''
\newblock in {\em Proc. Int. Conf. Learn. Represent.}, 2015, vol.~5.

\end{thebibliography}


\begin{thebibliography}{10}

\bibitem{tankus2001convexity}
Ariel Tankus and Yehezkel Yeshurun,
\newblock ``Convexity-based visual camouflage breaking,''
\newblock {\em Comput. Vis. Image Underst.}, vol. 82, pp. 208--237, 2001.

\bibitem{huerta2007improving}
Iv{\'a}n Huerta, Daniel Rowe, Mikhail Mozerov, and Jordi Gonz{\`a}lez,
\newblock ``Improving background subtraction based on a casuistry of colour-motion segmentation problems,''
\newblock in {\em Proc. Iberian Conf. Pattern Recognit. Image Anal.}, 2007, pp. 475--482.

\bibitem{sengottuvelan2008performance}
P~Sengottuvelan, Amitabh Wahi, and A~Shanmugam,
\newblock ``Performance of decamouflaging through exploratory image analysis,''
\newblock in {\em Proc. Int. Conf. Emerg. Trends Eng. Technol.}, 2008, pp. 6--10.

\bibitem{zeng2024mgqformer}
Kunlun Zeng, Ri~Cheng, Weimin Tan, and Bo~Yan,
\newblock ``Mgqformer: Mask-guided query-based transformer for image manipulation localization,''
\newblock in {\em Proceedings of the AAAI Conference on Artificial Intelligence}, 2024, vol.~38, pp. 6944--6952.

\bibitem{zhang2025excitation}
Lu~Zhang, Ri~Cheng, Zuyang He, Mei Mei, Bin Wu, Weimin Tan, Bo~Yan, Shangfeng Wang, and Fan Zhang,
\newblock ``Excitation-encoded single-emission shortwave infrared lanthanide fluorophore palette for real-time in vivo multispectral imaging,''
\newblock {\em Nature Photonics}, vol. 19, no. 11, pp. 1209--1218, 2025.

\bibitem{le2019anabranch}
Trung-Nghia Le, Tam~V Nguyen, Zhongliang Nie, Minh-Triet Tran, and Akihiro Sugimoto,
\newblock ``Anabranch network for camouflaged object segmentation,''
\newblock {\em Comput. Vis. Image Underst.}, vol. 184, pp. 45--56, 2019.

\bibitem{fan2020camouflaged}
Deng-Ping Fan, Ge-Peng Ji, Guolei Sun, Ming-Ming Cheng, Jianbing Shen, and Ling Shao,
\newblock ``Camouflaged object detection,''
\newblock in {\em Proc. IEEE/CVF Conf. Comput. Vis. Pattern Recognit.}, 2020, pp. 2777--2787.

\bibitem{lv2021simultaneously}
Yunqiu Lv, Jing Zhang, Yuchao Dai, Aixuan Li, Bowen Liu, Nick Barnes, and Deng-Ping Fan,
\newblock ``Simultaneously localize, segment and rank the camouflaged objects,''
\newblock in {\em Proc. IEEE/CVF Conf. Comput. Vis. Pattern Recognit.}, 2021, pp. 11591--11601.

\bibitem{sun2022boundary}
Yujia Sun, Shuo Wang, Chenglizhao Chen, and Tian-Zhu Xiang,
\newblock ``Boundary-guided camouflaged object detection,''
\newblock in {\em Proc. Int. Joint Conf. Artif. Intell.}, 2022, pp. 1335--1341.

\bibitem{pang2022zoom}
Youwei Pang, Xiaoqi Zhao, Tian-Zhu Xiang, Lihe Zhang, and Huchuan Lu,
\newblock ``Zoom in and out: A mixed-scale triplet network for camouflaged object detection,''
\newblock in {\em Proc. IEEE/CVF Conf. Comput. Vis. Pattern Recognit.}, 2022, pp. 2160--2170.

\bibitem{Ren_2025_ICCV}
Guangyu Ren, Hengyan Liu, Michalis Lazarou, and Tania Stathaki,
\newblock ``Multi-modal segment anything model for camouflaged scene segmentation,''
\newblock in {\em Proc. IEEE/CVF Int. Conf. Comput. Vis.}, 2025, pp. 19882--19892.

\bibitem{gao2025cod}
Dongyang Gao, Yichao Zhou, Hui Yan, Chen Chen, and Xiyuan Hu,
\newblock ``{COD-SAM}: Camouflage object detection using {SAM},''
\newblock {\em Pattern Recognit.}, p. 111826, 2025.

\bibitem{wang2023depth}
Qingwei Wang, Jinyu Yang, Xiaosheng Yu, Fangyi Wang, Peng Chen, and Feng Zheng,
\newblock ``Depth-aided camouflaged object detection,''
\newblock in {\em Proc. ACM Int. Conf. Multimedia}, 2023, pp. 3297--3306.

\bibitem{liu2024depth}
Xinran Liu, Lin Qi, Yuxuan Song, and Qi~Wen,
\newblock ``Depth awakens: A depth-perceptual attention fusion network for rgb-d camouflaged object detection,''
\newblock vol. 143, pp. 104924, 2024.

\bibitem{wang2024depth}
Liqiong Wang, Jinyu Yang, Yanfu Zhang, Fangyi Wang, and Feng Zheng,
\newblock ``Depth-aware concealed crop detection in dense agricultural scenes,''
\newblock in {\em Proc. IEEE/CVF Conf. Comput. Vis. Pattern Recognit.}, 2024, pp. 17201--17211.

\bibitem{de2005tutorial}
Pieter-Tjerk De~Boer, Dirk~P Kroese, Shie Mannor, and Reuven~Y Rubinstein,
\newblock ``A tutorial on the cross-entropy method,''
\newblock vol. 134, pp. 19--67, 2005.

\bibitem{mattyus2017deeproadmapper}
Gell{\'e}rt M{\'a}ttyus, Wenjie Luo, and Raquel Urtasun,
\newblock ``{DeepRoadMapper}: Extracting road topology from aerial images,''
\newblock in {\em Proc. IEEE/CVF Int. Conf. Comput. Vis.}, 2017, pp. 3438--3446.

\bibitem{margolin2014evaluate}
Ran Margolin, Lihi Zelnik-Manor, and Ayellet Tal,
\newblock ``How to evaluate foreground maps?,''
\newblock in {\em Proc. IEEE/CVF Conf. Comput. Vis. Pattern Recognit.}, 2014, pp. 248--255.

\bibitem{fan2018enhanced}
Deng-Ping Fan, Cheng Gong, Yang Cao, Bo~Ren, Ming-Ming Cheng, and Ali Borji,
\newblock ``Enhanced-alignment measure for binary foreground map evaluation,''
\newblock {\em arXiv preprint arXiv:1805.10421}, 2018.

\bibitem{fan2017structure}
Deng-Ping Fan, Ming-Ming Cheng, Yun Liu, Tao Li, and Ali Borji,
\newblock ``Structure-measure: A new way to evaluate foreground maps,''
\newblock in {\em Proc. IEEE/CVF Int. Conf. Comput. Vis.}, 2017, pp. 4548--4557.

\bibitem{skurowski2018animal}
Przemys{\l}aw Skurowski, Hassan Abdulameer, Jakub B{\l}aszczyk, Tomasz Depta, Adam Kornacki, and Przemys{\l}aw Kozie{\l},
\newblock ``Animal camouflage analysis: Chameleon database,'' 2018,
\newblock Unpublished Manuscript.

\bibitem{he2025distill}
Xiankang He, Dongyan Guo, Hongji Li, Ruibo Li, Ying Cui, and Chi Zhang,
\newblock ``Distill any depth: Distillation creates a stronger monocular depth estimator,''
\newblock {\em arXiv preprint arXiv: 2502.19204}, 2025.

\bibitem{yang2024depth}
Lihe Yang, Bingyi Kang, Zilong Huang, Xiaogang Xu, Jiashi Feng, and Hengshuang Zhao,
\newblock ``Depth anything: Unleashing the power of large-scale unlabeled data,''
\newblock in {\em Proceedings of the IEEE/CVF conference on computer vision and pattern recognition}, 2024, pp. 10371--10381.

\bibitem{yang2024depth2}
Lihe Yang, Bingyi Kang, Zilong Huang, Zhen Zhao, Xiaogang Xu, Jiashi Feng, and Hengshuang Zhao,
\newblock ``Depth anything v2,''
\newblock {\em Advances in Neural Information Processing Systems}, vol. 37, pp. 21875--21911, 2024.

\bibitem{huang2023feature}
Zhou Huang, Hang Dai, Tian-Zhu Xiang, Shuo Wang, Huai-Xin Chen, Jie Qin, and Huan Xiong,
\newblock ``Feature shrinkage pyramid for camouflaged object detection with transformers,''
\newblock in {\em Proc. IEEE/CVF Conf. Comput. Vis. Pattern Recognit.}, 2023, pp. 5557--5566.

\bibitem{hao2025simple}
Chao Hao, Zitong Yu, Xin Liu, Jun Xu, Huanjing Yue, and Jingyu Yang,
\newblock ``A simple yet effective network based on vision transformer for camouflaged object and salient object detection,''
\newblock {\em IEEE Trans. Image Process.}, vol. 34, pp. 608--622, 2025.

\end{thebibliography}
\end{document}